\documentclass[letterpaper]{article} 
\usepackage{booktabs}

\usepackage[preprint]{rlj} 

\usepackage{amssymb}            
\usepackage{mathtools}          
\usepackage{mathrsfs}           
\usepackage{graphicx}           
\usepackage{subcaption}         
\usepackage[space]{grffile}     
\usepackage{url}                
\usepackage{lipsum}             
\usepackage{float}

\newtheorem{definition}{Definition}

\title{Frequency and Generalization of Periodic Activation Functions in Reinforcement Learning}

\setrunningtitle{Frequency and Generalization of Periodic Activation Functions}

\author{Augustine N. Mavor-Parker\textsuperscript{1$\dagger$}, Matthew J. Sargent\textsuperscript{1}, Caswell Barry\textsuperscript{2}, Lewis D. Griffin\textsuperscript{1}, Clare Lyle\textsuperscript{3}}

\emails{\{a.mavor-parker,m.sargent,l.griffin\}@cs.ucl.ac.uk \\
caswell.barry@ucl.ac.uk\\
clarelyle@deepmind.com}

\affiliations{
$^{1}$\textbf{Department of Computer Science, University College London}\\
$^{2}$\textbf{Department of Cell and Developmental Biology, University College London}\\
$^{3}$\textbf{Google Deepmind}\\
\par 
$^\dagger$ Work partially done while visiting the OATML group at Oxford University where Clare Lyle was a PhD student.
}

\contribution{
    We show that recent periodic activation function architectures trained on continuous control tasks consistently converge to similar frequencies regardless of their initialization frequency, unless frequencies are initialized excessively high.
    }
    {
    Previous works highlight the benefits of either low \citep{li2021functional} or high \citep{yang2022overcoming} frequency representations learned by different architectures. We find all architectures converge to a similar frequency, which we characterize in contribution 2.
    }

\contribution{
    We demonstrate that representations learned with periodic activation functions do not outperform ReLU activations when Gaussian noise is added to network inputs, suggesting that learned Fourier features are more prone to overfitting than ReLU features and hence can be considered \textit{high frequency}.
    }
    {
    Prior work \citep{li2021functional, yang2022overcoming} established that periodic activation functions outperform ReLU activation functions by increasing sample efficiency but provided potentially incompatible descriptions of representation frequency and did not analyze the generalization properties of periodic representations.
    }

\contribution{
    We show that in some environments weight decay can partially offset the poor robustness of Fourier features by punishing excessive frequency growth during training.
    }
    {
    This is slightly surprising, previous works have shown that regularization techniques used in traditional deep learning often do not benefit deep reinforcement learning agents (see \citet{gogianu2021spectral}).
    }


\summary{Periodic activation functions, often referred to as \textit{learned Fourier features} have been demonstrated to improve the sample efficiency and stability of deep RL algorithms. A critical property of learned Fourier features is their frequency, which describes how quickly features oscillate as network inputs change. Ostensibly incompatible hypotheses have been made about why Fourier feature representations improve sample efficiency and how the frequencies of Fourier features contributes to such improvements. One hypothesis is that periodic activations learn \textit{low frequency} representations and as a result avoid overfitting to bootstrapped targets. Another hypothesis is that periodic activations learn \textit{high frequency} representations that are more expressive, allowing networks to quickly fit complex value functions. 

We analyze these claims empirically, finding that different Fourier feature architectures see their frequencies increase substantially throughout training regardless of their initialization frequency. To characterize the frequencies of Fourier representations, we first analyze how they perform when noise perturbations are added to their inputs, showing that they exhibit worse robustness when compared to otherwise identical networks with ReLU activation functions. Then, we compute representation properties that measure expressivity, such as effective rank and the cosine similarity between representations before and after non-linearities. From these empirical results, we conclude that Fourier feature architectures trained with adaptive optimizers generally lead to \textit{high frequency} representations when compared to otherwise identical ReLU representations. Finally, we show that weight decay regularization is able to partially offset the overfitting of periodic activation functions, delivering value functions that can learn quickly while also being robust to input noise.
}

\begin{document}

\maketitle  

\begin{abstract}
Periodic activation functions, often referred to as \textit{learned Fourier features} have been demonstrated to improve the sample efficiency and stability of deep RL algorithms. Ostensibly incompatible hypotheses have been made about the source of these improvements. One is that periodic activations learn \textit{low frequency} representations and as a result avoid overfitting to bootstrapped targets. Another is that periodic activations learn \textit{high frequency} representations that are more expressive, allowing networks to quickly fit complex value functions. We analyze these claims empirically, finding that periodic activation function architectures consistently converge to high frequency representations regardless of their initialization frequencies. As a result, we also find that while periodic activation functions improve sample efficiency, they exhibit worse generalization on states with added observation noise---especially when compared to otherwise identical networks with ReLU activation functions. Finally, we show that weight decay regularization is able to partially offset the overfitting of periodic activation functions, delivering value functions that learn quickly while also generalizing.
\end{abstract}

\section{Introduction}
Deep learning has enabled reinforcement learning (RL) to conquer environments with large, high dimensional state spaces (e.g. \citet{silver2016mastering, badia2020agent57}). The architectures of deep RL agents have largely emulated innovations from supervised learning---a prime example being the original deep Q-networks \citep{silver2013playing}, which used the GPU-optimized implementation of convolutional layers from AlexNet \citep{krizhevsky2012imagenet}. However, design choices inherited from supervised learning do not always readily meet the demands of RL. For example, popular techniques such as weight decay~\citep{salimans2016weight} and momentum~\citep{bengio2021correcting} do not improve performance in RL as consistently as they do in the supervised learning contexts in which they were initially developed. 

One important dimension along which RL differs from supervised learning is in the trade-off between generalization and memorization, with RL algorithms often benefiting from localized features~\citep{ghiassian2020improving} such as tile coding~\citep{sherstov2005function}. Localized features limit how well a function approximator can generalize. While some degree of generalization is necessary for large problem scales, over generalizing can hinder agent performance---i.e. value targets in one state should not necessarily influence value predictions in a distant unrelated state. \textit{Fourier features}, a popular representation for both supervised learning~\citep{rahimi2007random} as well as linear RL~\citep{konidaris2011value}, present a means of balancing generalization and memorization, whereby the degree of generalization between training samples of the ensuing linear function approximator can be precisely specified by the \textit{frequency} with which these features are initialized. A Fourier feature's frequency defines how quickly a feature oscillates as input values change---frequency is closely related to the notion of value function smoothness introduced by \citet{lyle2022learning}, which measures the magnitude of the change in predicted values between two consecutive environment states. Low frequencies bias the learner towards smooth functions that generalize more between training samples, whereas high-frequencies will bias the learner towards non-smooth functions.
\begin{figure}
\centering
\begin{subfigure}{0.4\textwidth}
    \centering
    \includegraphics[width=.99\linewidth]{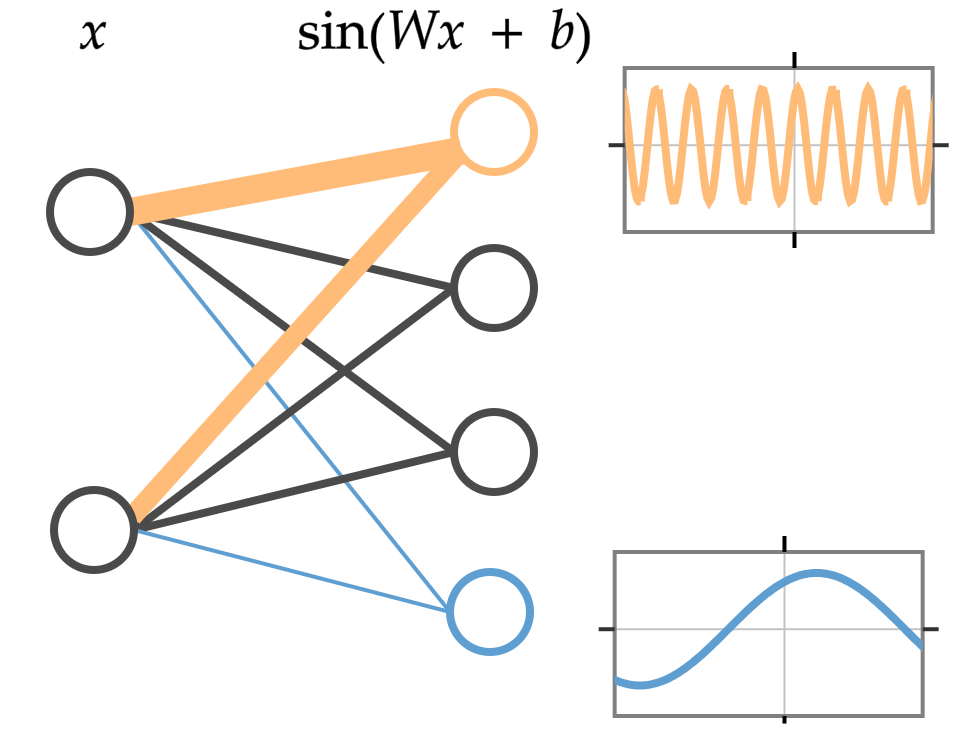}
    \caption{}
    \label{fig:LFF_architecture}
\end{subfigure}
\hspace{1.2cm}
\begin{subfigure}{0.39\textwidth}
    \centering
    \includegraphics[width=.99\linewidth]{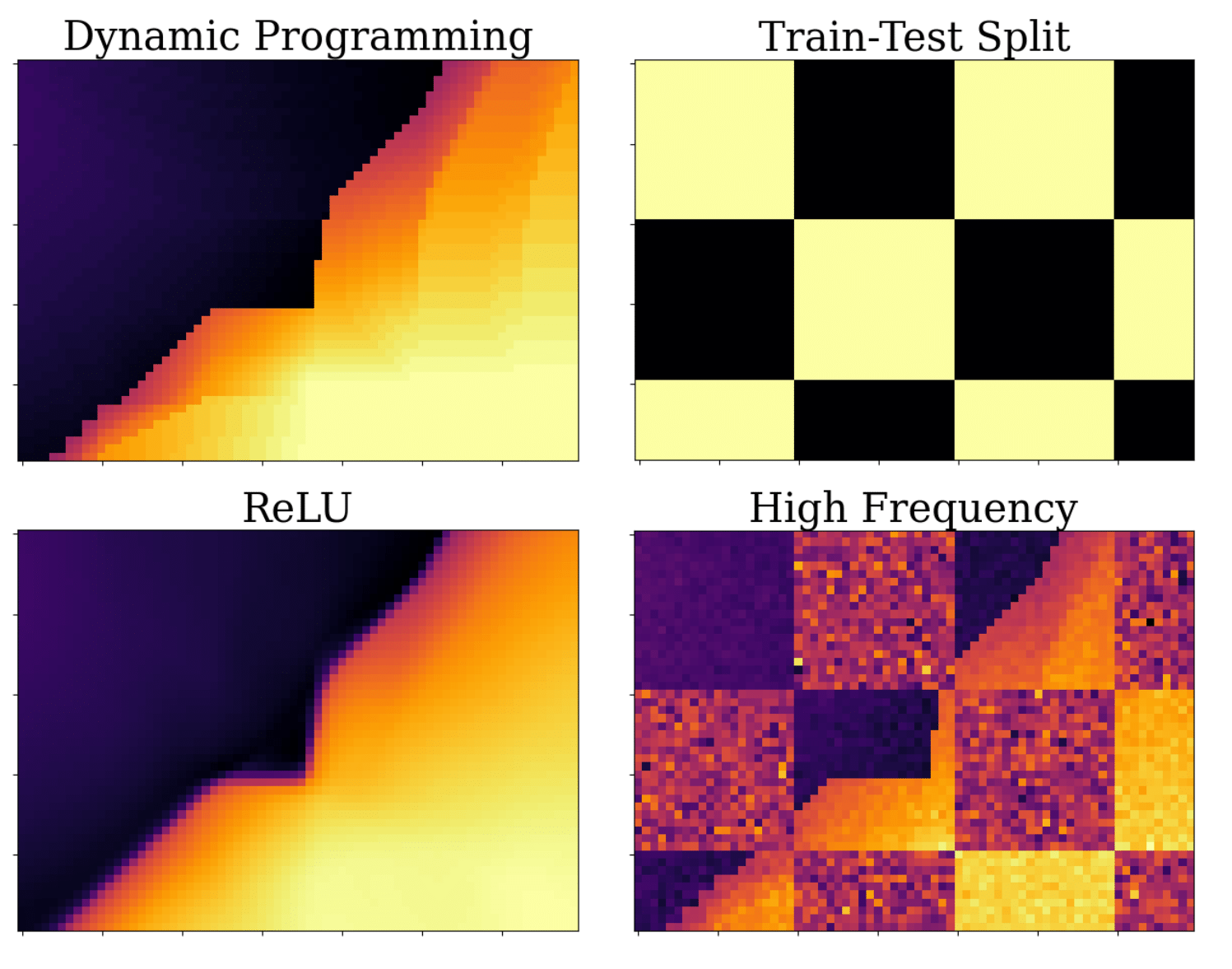}
    \caption{}
    \label{fig:LFF_motivation}
\end{subfigure}
\caption{(a) Learned Fourier feature layers are fully-connected layers with a $\sin$ activation. Larger weight magnitudes lead to faster oscillations as shown by the orange line, while smaller weights lead to slower oscillations as shown by the blue lines. (b) Results of supervised learning warm up experiment using high frequency Fourier features as well as ReLU features to fit the dynamic programming ground truth of the mountain car value function \citep{moore1990efficient}. Top left shows a segment of the ground truth value function from the mountain car environment. Top right shows the training/test distributions used by the ReLU and high frequency Fourier feature architectures: light is the training set, dark is the test set. The ReLU network smoothly fits the training regions and generalizes well to the test regions (bottom left). High frequency Fourier features fit the training distribution more precisely but do not generalize to the test regions (bottom right). See supplementary \ref{app:supervised_learning_warmup} for details.}
\label{fig:supervised_value_functions}
\end{figure}

Fourier features initially appeared in the reinforcement learning literature as fixed Fourier decompositions of network inputs, which were then fed to linear function approximators \citep{konidaris2011value}. This approach is difficult to scale to high dimensional inputs as the number of features increases rapidly with the size of the input (see \citet{brellmann2023fourier} for work on scaling these traditional Fourier features to high dimensional deep RL tasks). In recent years, an alternative approach for constructing Fourier features has emerged from the computer vision literature: learned Fourier features. Learned Fourier feature architectures replace typical deep learning activation functions like ReLUs with sinusoidal activations~\citep{NEURIPS2020_53c04118}. These periodic activation functions replace the fixed frequencies of \citet{rahimi2007random} and \citet{konidaris2011value} with feature frequencies that are learned from experience and determined by networks weights (see figure \ref{fig:LFF_architecture}). 

\citet{li2021functional} and \citet{yang2022overcoming} applied periodic activations to deep RL, incorporating them into a neural network architecture trained to perform value function approximation.
\citet{yang2022overcoming} argue that the use of a periodic activation functions improves sample efficiency because it causes networks to learn \textit{high frequency} representations that quickly fit high frequency details in value functions. Concurrently, \cite{li2021functional} also employed periodic activations but with a different perspective. \citet{li2021functional} state that periodic activations learn \textit{low frequency} representations, which improves sample efficiency by reducing the amount of overfitting to noisy targets. It is not clear why both works see similar improvements on the same benchmark environments when \cite{yang2022overcoming} suggests that high frequency representations are useful while \cite{li2021functional} argues that low frequencies are beneficial.   
\newline
\newline
We investigate how previous works with similar methodologies can arrive at these different conclusions. We are motivated by a simple question: do learned Fourier features help in deep RL because they learn high-frequency representations that are more expressive than ReLU representations, or because they learn low-frequency representations than enable better generalization and smoother bootstrapping targets? 
Our findings are summarized as follows: first, we observe the frequency of representations throughout training, finding that the both the architecture from \citet{li2021functional} and \citet{yang2022overcoming} converge to similar frequencies, that are significantly larger than their initialization frequencies. This growth in frequency can largely be attributed to the growth of the network's parameter norm, a widely-observed phenomenon in deep reinforcement learning \citep{nikishin2022primacy, dohare2023loss}. In section \ref{sec:generalization}, we find that the improvements of learned Fourier features over ReLU features are less apparent when noise is added to network inputs---suggesting that learned Features are not as smooth \citep{lyle2022learning} as ReLU features, and hence can be characterized as \textit{high frequency}. In section \ref{sec:understanding_bad_generalization} we connect feature frequency with policy instability, finding that the learned Fourier features are much more sensitive to changes in their input than ReLU features and exhibit a higher effective rank. In section \ref{sec:fixing_generalization}, we show that weight decay regularization partially impedes runaway frequency growth, allowing Fourier features to learn quickly but also be robust to perturbations in input observations.

\section{Related Work}
Periodic representations are widely applicable across different domains of machine learning research. In computer vision \citep{NEURIPS2020_53c04118, tancik2020fourier} Fourier-like positional encodings are used when constructing 3D scenes from 2D images \citep{mildenhall2021nerf}. For solving partial differential equations, neural Fourier operators learn representations that operate at a range of resolutions \citep{li2020fourier}. Additionally, geometric deep learning often operates in Fourier space \citep{cohen2018spherical, cobb2020efficient}. \citet{rahimi2007random} show that random Fourier features can provide a useful approximation for kernel methods in supervised learning. 
Even when not explicitly engineered into an architecture, Fourier features have emerged in transformer architectures \citep{nanda2023progress}.

In RL, the use of Fourier features was initiated by \citet{konidaris2011value}, who learn a value function on top of a fixed Fourier decomposition of each individual variable in the state space. These classic Fourier features has proven to be useful for generalization in spatial navigation tasks \citep{yu2020prediction}. Recently, \citet{li2021functional} and \citet{yang2022overcoming} developed a scalable approach for learning periodic features---achieving state of the art results at the time of publication.  

One potential drawback of Fourier representations is that they continue to oscillate outside of their training distribution, leading to inaccurate extrapolation behaviour \citep{beukman2022adaptive}. More localized feature representations such as tile coding \citep{sherstov2005function} can avoid this extrapolation behaviour. \citet{ghiassian2020improving} leveraged localized feature representations to demonstrate the importance of fitting discontinuous local components of the value function. While localized representations have found some success ~\citep{whiteson2010adaptive}, they have not seen widespread adoption into neural network architectures to the same degree as periodic activations, whose use in neural networks goes back several decades~\citep{mccaughan1997on}. 

Generalization in deep reinforcement learning is a more nuanced property than in supervised learning due to its effect on the stability of RL algorithms \citep{van2018deep}. Generalization between observations in a single environment is a double-edged sword, bringing the potential of accelerating gradient-based optimization methods~\citep{jacot2018neural} but also the risk of over-estimation and divergence of the bootstrapped training objective~\citep{achiam2019towards}. Many works studying generalization in deep reinforcement learning focus on out-of-distribution transfer to new tasks, either drawn from some independent and identically distributed distribution~\citep{cobbe2019quantifying}, or obtained by changing the reward function~\citep{dayan1993improving, kulkarni2016deep} or some latent factor of the environment transition dynamics. For a comprehensive review of generalization in reinforcement learning, we refer to the work of \citet{kirk2023survey}. Our interest will be in the relationship between low-frequency representations and generalization within a single environment, a connection previously studied in the RL context by \citet{lyle2022learning}. 

\section{Reinforcement Learning with Periodic Activations}\label{sec:method}
Reinforcement learning is formalised in the framework of Markov decision processes, which are a tuple $\mathcal{M}={\langle \mathcal{S}, \mathcal{A}, \mathcal{R}, \mathcal{P}, \gamma \rangle}$ where $\mathcal{S}$ is the set of all states an agent can experience; $\mathcal{A}$ is the set of actions an agent can take; $\mathcal{R}: \mathcal{S}\times\mathcal{A}\rightarrow\mathbb{R}$ is a reward function that gives rewards to the agent; $\gamma \in [0, 1]$ is the discount factor that controls how preferential near term rewards are to long term rewards and $\mathcal{P}: \mathcal{S}\times\mathcal{R}\times\mathcal{S}\times\mathcal{A}\rightarrow[0, 1]$ is the transition function, where $\mathcal{P}(s', r, s, a)=P(S_{t}=s',R_{t}=r|S_{t-1}=s,A_{t-1}=a)$, for $S_{t} \in \mathcal{S}$, $A_{t} \in \mathcal{A}$ and $t$ indexes the timestep \cite[p. 47]{sutton2018reinforcement}. This work focuses on episodic reinforcement learning where  an agent's goal is to maximize its expected return $G=\mathbb{E}_{\pi}[\sum_{k=0}^{T}\gamma^{k} r_{t+k+1}]$ in a given episode, where $T$ is the length of the episode and $\pi: \mathcal{S} \rightarrow \mathcal{A}$ is a policy that controls how the agent selects its actions.

Sample efficient neural networks for continuous control are off-policy and actor-critic based, meaning they learn a policy function and an action-value function. Action-value functions $Q: \mathcal{S} \times \mathcal{A} \rightarrow \mathbb{R}$ predict the expected future return given the current state and action $Q(s, a)=\mathbb{E}_{\pi}\left[G_{t}|S_{t}=s, A_{t}=a\right]$ \cite[p. 58]{sutton2018reinforcement}. Learned Fourier features have been shown to be most useful when used within the value functions of actor-critic architectures \citep{li2021functional, yang2022overcoming}, and hence we focus our experiments on value learning.

\subsection{Representation Frequency}
Throughout this work we describe representations as having a frequency. To avoid ambiguity we provide a a description of the frequency of representations here.
\begin{definition}\label{def:representation_frequency}
Given a dataset of network inputs $\mathcal{D}$ and layer weights $\mathbf{W}$, the representation frequency $f$ of the $i^{th}$ neuron is the maximum activation minus the minimum activation, divided by $2\pi$. This definition corresponds to the number of wavelengths covered by the $i^{th}$ activation over the dataset $\mathcal{D}$.
\begin{equation}
    f_{i}= \left[\frac{\max_{\mathbf{x}\in\mathcal{D}}{\left[(\mathbf{W}\mathbf{x})_{i}\right]} - \min_{\mathbf{x}\in\mathcal{D}}\left[({\mathbf{W}\mathbf{x}})_{i}\right]}{2\pi}\right]
\end{equation} 
\end{definition}
Intuitively, one can plot the activation of a given neuron for different network inputs, which traces out either a sine or cosine wave depending on the choice of periodic activation function. As weight values become larger, the number of periods covered a periodic activation increases. Our definition follow this intuition, counting how many wavelengths the activation oscillates through in the over a dataset of inputs.

Next, we describe the two different flavours of learned Fourier feature architectures from the literature that we call \textit{learned Fourier features} \citep{yang2022overcoming} and \textit{concatenated learned Fourier features} \citep{li2021functional}. It is important to note these are equivalent to traditional deep RL architectures, except that they replace activation functions with a sinusoid and/or cosine activation. They do not perform a Fourier decomposition of each state variable as is done by \citep{konidaris2011value}.
\subsection{Learned Fourier Features (LFF)}\label{sec:learned_fourier_features}
The learned Fourier features (LFF) of \cite{yang2022overcoming} learn a value function that has a multi-layer perceptron architecture. The first layer of learned Fourier feature networks take the following form $\mathcal{F}=\sin(\mathbf{W} \mathbf{x} + \mathbf{b})$, where $\mathbf{W}$ and $\mathbf{b}$ are weights and biases and $\mathbf{x}$ is the layer input. Then, layers of additional weights and biases are stacked on top of the initial Fourier layer---these subsequent layers have ReLU activations. We denote the ReLU layers as $\mathcal{L}$, meaning an $N$-layer learned Fourier feature value function be written as $Q(s, a)=\mathcal{L}_{N}(\mathcal{L}_{N-1}(... (\mathcal{F}_{1}(s, a)))$, where the state vector $s$ and the action vector $a$ are concatenated into one large vector $x$, which is the input to the learned Fourier feature layer (using a notation similar to \citep{li2021functional}). \cite{yang2022overcoming} initializes weights from a normal distribution and their biases from a uniform distribution, which using their notation, is formulated as the following.
\newline
\newline
\begin{minipage}{0.45\textwidth}
    \begin{equation}\label{eqn:LFF_initialisation}
    \mathbf{W}_{i,j} \sim \mathcal{N}(0, \frac{\pi \beta}{d})
    \end{equation}
\end{minipage}
\begin{minipage}{0.45\textwidth}
    \begin{equation}
    \mathbf{b}_{j} \sim \mathcal{U}(-\pi, \pi)
    \end{equation}
\end{minipage}
\newline
\newline
Where $\beta$ is the initial bandwidth and $d$ is the dimension of the layer input. Larger weight values $\mathbf{W}_{ij}$ mean higher frequency oscillations as the network input changes, while smaller weights mean lower frequency oscillations. The bias $\mathbf{b}_{j}$ can be thought of as controlling the phase of the output neurons in the first layer, translating the representation along the output oscillations of each individual neuron. \cite{yang2022overcoming} hypothesise that this architecture allows networks to learn high frequency representations, which prevents the underfitting of value functions.
\subsection{Concatenated Learned Fourier Features (CLFF)}\label{sec:concatenated_learned_fourier_features}
\cite{li2021functional} developed concatenated learned Fourier features (CLFF), which concatenate the network inputs alongside periodic representations in the first layer of their networks. In addition, \cite{li2021functional} use both cosine and sine activations and concatenate these features into one large vector (which additionally includes the raw network input).
\begin{equation}\label{eqn:CLFF_initialisation}
    \mathcal{CF}=( \sin(\mathbf{W}\mathbf{x}), \cos(\mathbf{W}\mathbf{x}), \mathbf{x})
\end{equation}
Where the comma denotes concatenation of $\sin(\mathbf{W}\mathbf{x})$, $\cos(\mathbf{W}\mathbf{x})$ and $\mathbf{x}$ into one larger vector.  Note that \cite{li2021functional} do not include a bias term in their layer architecture. The rest of the architecture follows the same pattern as learned Fourier features with ReLU layers following the initial layer as follows: $Q(s, a)=\mathcal{L}_{N}(\mathcal{L}_{N-1}(... (\mathcal{CF}_{1}(s, a)))$. The motivation of \cite{li2021functional} was to avoid overfitting to noise in bootstrapped targets when fitting the value function. It was hypothesized that \textit{the periodic portion of the representation would learn smooth representations, while the concatenated network input ($x$) would retain the high frequency information for downstream layers}.
\section{Experiments}
Here we empirically investigate the representations of periodic activation functions introduced in section \ref{sec:method} focusing on Deepmind control \citep{tassa2018deepmind}, a standard benchmark for continuous control consisting of a variety of agent morphologies (quadrupeds, fingers and humanoids) paired with reward functions that encourage skills like running or walking. We focus on control from proprioceptive observations. The action space is a continuous vector that controls the forces to applied to different joints. The full set of environments and hyperparameters used can be found in supplementary \ref{app:deepmind_hyperparams}. Our code builds upon the \textit{jaxrl} repository \citep{jax2018github, jaxrl} with reference to the original repositories of \citet{li2021functional} and \citet{yang2022overcoming}.

\citet{li2021functional} and \citet{yang2022overcoming} agree that periodic activation functions are most useful for off-policy algorithms. Furthermore, ablations have shown periodic activations are only beneficial when added to the critic function of actor-critic architectures \citet{li2021functional, yang2022overcoming}. As a result, we only experiment with periodic activation functions embedded within critic networks, which themselves are embedded in the (off-policy) soft-actor critic learning algorithm \citep{haarnoja2018soft}.

Following \citet{yang2022overcoming}, we use a learned Fourier feature critic with a first layer width equal to 40 times the size of the input, followed by two ReLU activated hidden layers of width 1024. The concatenated learned Fourier features critic has an initial layer with the same total width. The ReLU critic has an equivalent architecture except with a ReLU used instead of a periodic activation. All agents have the same actor architecture (details in supplementary \ref{app:relu_architecture}). We use five seeds for each data point in section \ref{sec:measuring_frequency} and ten seeds in the proceeding sections. Error bars and shaded regions indicate the standard deviation throughout.


\subsection{Representation Frequency and In-Distribution Performance}\label{sec:measuring_frequency}
\begin{figure}[H]
\centering
\begin{subfigure}{0.31\textwidth}
    \centering
    \includegraphics[width=.99\linewidth]{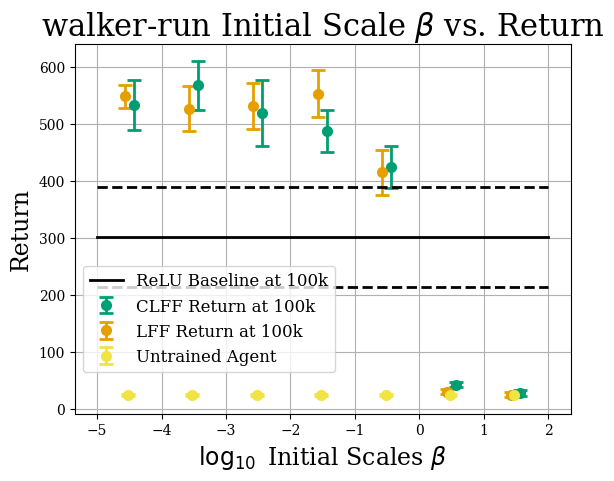}
    \caption{}
    \label{fig:scales_vs_return}
\end{subfigure}
\begin{subfigure}{0.31\textwidth}
    \centering
    \includegraphics[width=.99\textwidth]{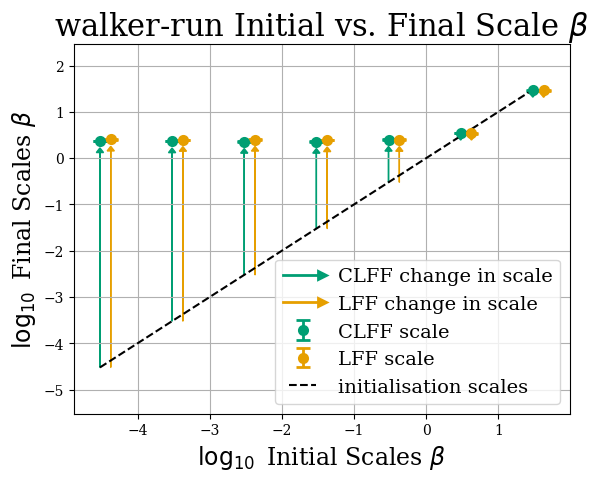}
    \caption{}
    \label{fig:scales_before_and_after_training}
\end{subfigure}
\begin{subfigure}{0.32\textwidth}
    \centering
    \includegraphics[width=.99\linewidth]{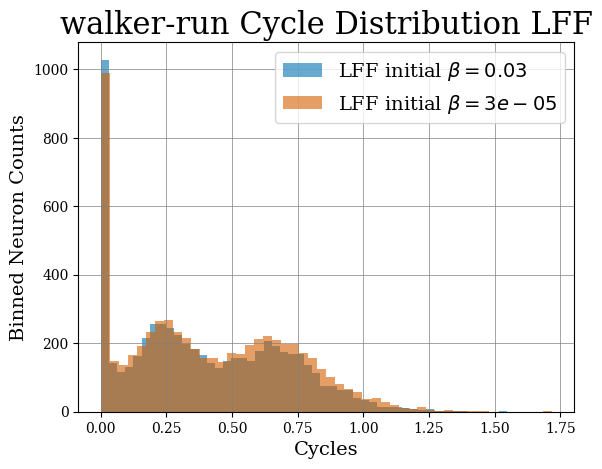}
    \caption{}
    \label{fig:walker_run_cycle_distribution}
\end{subfigure}
\caption{Regardless of initialization and architecture (LFF vs CLFF), Fourier features converge to similar frequencies. (a) shows return early in training with different initial $\beta$'s, demonstrating that periodic activation functions improve performance at a range of initialization frequencies. (b) shows that the final scale $\beta$ of learned Fourier features (i.e. the frequency learned) is similar regardless of the initialization frequency. (c) shows the distribution of cycles (which is proportional to $\beta$) for different initial $\beta$'s. Error bars represent standard deviation over five seeds in panels (a) and (b).}
\end{figure}

\begin{tcolorbox}[colframe=black!50, colback=black!5, rounded corners]
We reproduce the results of \citet{li2021functional} and \citet{yang2022overcoming}, showing that periodic activation functions either improve or do not hinder returns received early in training for most initialization frequencies (i.e. they increase sample efficiency), provided the initialization frequency is not excessively high.
\end{tcolorbox}
We analyze representation frequency on three environments highlighted by \citet{yang2022overcoming}: hopper-hop, walker-run and quadruped-run (we only focus on 3 environments initially due to computational limits, in later sections we broaden our analysis to all 8 environments tested by \citet{yang2022overcoming}). We reproduce the results of both \citet{yang2022overcoming} and \citet{li2021functional}, showing that both approaches can be more sample efficient than ReLU architectures on some environments. Shown in figure \ref{fig:scales_vs_return} are the returns for walker-run at 100k environment steps (roughly when the number of timesteps when improved performance of learned Fourier is most apparent), appendix \ref{app:measuring_frequencies} shows results for quadruped-run (where LFF and CLFF outperform ReLU) and hopper-hop (where LFF and CLFF match ReLU). We plot performance over a range of different initialization scales $\beta$ (see equation \ref{eqn:LFF_initialisation}). In addition to the results provided in the main section, in the supplementary we also reproduce the improvements of learned Fourier features across all 8 environments from the Deepmind control suite \ref{fig:no_noise_return_curves}.
\label{fig:frequency_convergences}
For low initial scales, learned Fourier features and concatenated learned Fourier features outperform the ReLU baseline (in the low sample limit shown in figure \ref{fig:scales_vs_return}). The performance of the Fourier feature architectures is relatively consistent for low initial $\beta$. If very high scale initializations are used ($\beta>0.03$, roughly equal to $-0.5$ on the x-axis of figure \ref{fig:scales_vs_return} and figure \ref{fig:scales_before_and_after_training}), the performance of the periodic activations drop dramatically. Furthermore, the performance of learned Fourier features relative to concatenated learned Fourier features is roughly comparable---regardless of initialization scales. We measure the weight scales and performances for the two additional environments in appendix \ref{app:measuring_frequencies}.

\subsubsection{Representation Frequencies After Training}
\begin{tcolorbox}[colframe=black!50, colback=black!5, rounded corners]
We find that the different architectures of \citet{li2021functional} and \citet{yang2022overcoming} learn similar representation frequencies regardless of the frequency used for initialization, suggesting that initialization frequency and architecture choices do not significantly affect the ultimate frequency of periodic representations.
\end{tcolorbox}
Previously, it was hypothesized that lower initial $\beta$'s would lead to lower frequency representations after training, which could improve generalization \citep{li2021functional}. However, an explicit comparison between the frequencies learned by learned Fourier features and concatenated learned Fourier features was not made in previous works. We measure weight scales $\beta$ by assuming the weight distribution is a diagonal Gaussian, and then computing its standard deviation to extract $\beta$ (see equation \ref{eqn:LFF_initialisation}).

Weight distributions generally rise to similar scales after training regardless of their initialization. In figure \ref{fig:scales_before_and_after_training}, we find that despite differing initialization scales (shown by the black dotted line), weight distributions generally rise to similar scales (approximately $\beta\approx10^{0.3}\approx2$, which is roughly consistent with the standard deviations reported in \citet{li2021functional}, see supplementary \ref{app:li_pathak_std_dev} for details). This suggests that initialization scales are not as important as previously thought for determining final scales \citep{yang2022overcoming}. The exception being for large scale initializations, where it seems that if representation frequency is initialized too high, it is not able to accurately predict the values of state-action pairs (see the far right hand side of figure \ref{fig:scales_vs_return} and \ref{fig:scales_before_and_after_training}). It also suggests that the respective architecture choices of \citet{li2021functional} and \citet{yang2022overcoming} do not influence the ultimate frequencies of Fourier representations.

For a more intuitive measure of representation frequency, we count the number of wavelengths covered for each neuron for a given batch, which is equivalent to definition \ref{def:representation_frequency}. We plot the distribution of these frequencies for all neurons in figure \ref{fig:walker_run_cycle_distribution} for two different runs, a run with a low frequency initialization and a run with a high frequency initialization. The frequency distribution is strikingly similar for a large initial $\beta$ and a small initial $\beta$, further suggesting that initial scale is not consequential for the final scales learned periodic activations.

In summary, learned Fourier features and concatenated learned Fourier features generally rise to similar weight scales after training. So far, we are yet to concretely characterize the frequency of periodic activations when compared to more traditional architectures that employ ReLU activation functions. Crudely, one could consider $\frac{1}{4}$ of a sinusoidal cycle as roughly equivalent to the non-linearity of a ReLU function---which would suggest that the representation frequencies learned by learned Fourier feature architectures are mostly high frequency (see figure \ref{fig:walker_run_cycle_distribution}). Prior works suggest a connection between low frequency components of a function and generalization~\citep{rahaman2019on}. With this in mind, in the next section we develop a more principled notion of the frequency of learned Fourier features by measuring their generalization properties. 

    
\subsection{Generalization Properties of Learned Fourier Features}\label{sec:generalization}
\begin{tcolorbox}[colframe=black!50, colback=black!5, rounded corners]
Learned Fourier features no longer outperform ReLU features when Gaussian noise is added to network inputs, suggesting that learned Fourier features are not ``smooth'' \citet{lyle2022learning} and can be characterized as \textit{high frequency}.
\end{tcolorbox}
Following the literature on generalization in RL \citep{dulac2021challenges}, we benchmark generalization by perturbing state observations with Gaussian noise---we only perturb the observations at test time. No training is performed on observations with added noise. As we have shown learned Fourier features and concatenated learned Fourier features learn representations with similar frequencies, we focus our analysis in this section on the simpler learned Fourier features. We initialize the initial frequencies scales for learned Fourier features using the scales from \citet{yang2022overcoming}.
Noise levels were tuned by hand for each environment, as the environments have different observation distributions (see supplementary
\ref{tab:tuned_noise_levels_per_environment} for noise scales). We tuned observation noise to 3 different levels per environment: low noise where both ReLU and Fourier features can deploy their policy from training without degradation; medium noise where there is a noticeable impact on performance but the policies learned during training are still useful when compared to an untrained policy; and lastly high noise, where policy performance collapses.
\begin{figure}[H]
\centering
\begin{subfigure}{0.33\textwidth}
    \centering
    \includegraphics[width=.99\linewidth]{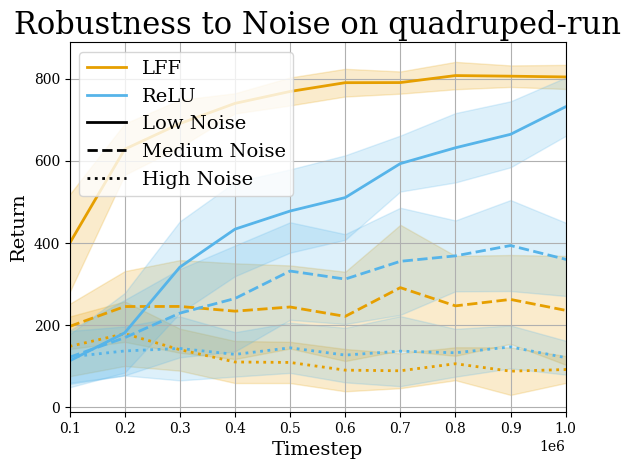}
    \caption{}
    \label{fig:quadruped_generalisation}
\end{subfigure}
\begin{subfigure}{0.31\textwidth}
    \centering
    \includegraphics[width=.99\linewidth]{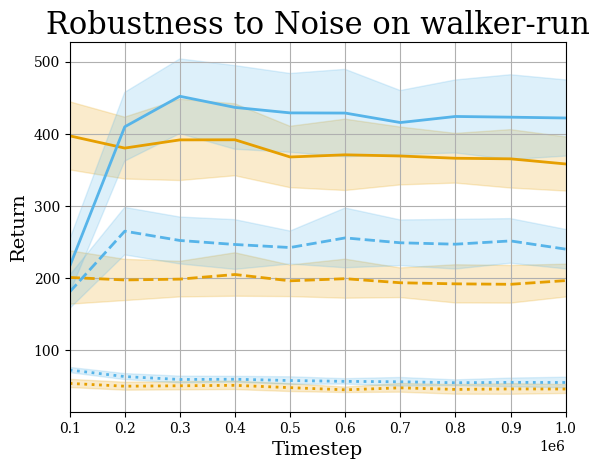}
    \caption{}
    \label{SUBFIGURE LABEL 3}
\end{subfigure}
\begin{subfigure}{0.31\textwidth}
    \centering
    \includegraphics[width=.99\textwidth]{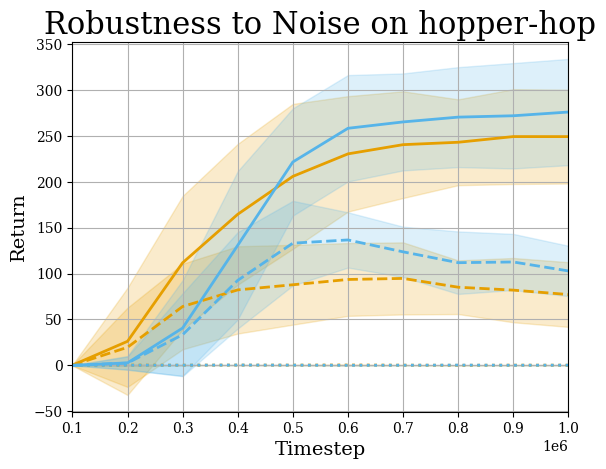}
    \caption{}
    \label{SUBFIGURE LABEL 1}
\end{subfigure}
\caption{Learned Fourier features perform either as good or worse than ReLU features when input observations are perturbed with medium noise. We apply three different levels of noise to observations at test time---in the medium noise case we find that architectures with learned Fourier feature activations generally either as good or sometimes worse (like in walker-run) than ReLU architectures. Results are reported across ten seeds, shaded region indicates standard deviation. Evaluation at different noise levels is performed at 100k increments throughout training.}
\end{figure}\label{fig:generalization}

In the medium noise regime, learned Fourier features are either worse than or comparable to ReLU activation functions at generalizing to noisy observations. This is in contrast to the results from previous work \citep{yang2022overcoming} with no noise, where learned Fourier features outperform ReLU in walker-run and quadruped-run (see supplementary figure \ref{fig:no_noise_return_curves}). Consider the quadruped-run results in figure \ref{fig:quadruped_generalisation}, the dotted blue line at 900k steps has a return of approximately 400, while the equivalent learned Fourier architecture has a return of approximately 250. This is despite the fact that at 900k timesteps, the performance of LFF with low noise, shown by the solid orange line, achieves a higher return when compared to the ReLU with low noise. These results suggest that periodic activation functions learn high frequency representations, as their improved performance when compared to ReLU activations disappears when Gaussian noise is added to network inputs. 
\subsection{Why do Fourier Representations Struggle to Generalize?}\label{sec:understanding_bad_generalization}
\begin{figure}[H]
\centering
\begin{subfigure}{0.34\textwidth}
    \centering
    \includegraphics[width=.99\linewidth]{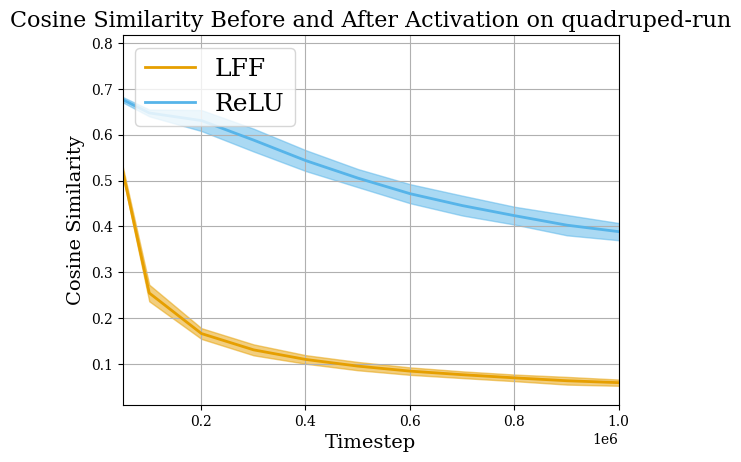}
    \caption{}
    \label{fig:cosine_similarity_after_noise}
\end{subfigure}
\begin{subfigure}{0.33\textwidth}
    \centering
    \includegraphics[width=.99\linewidth]{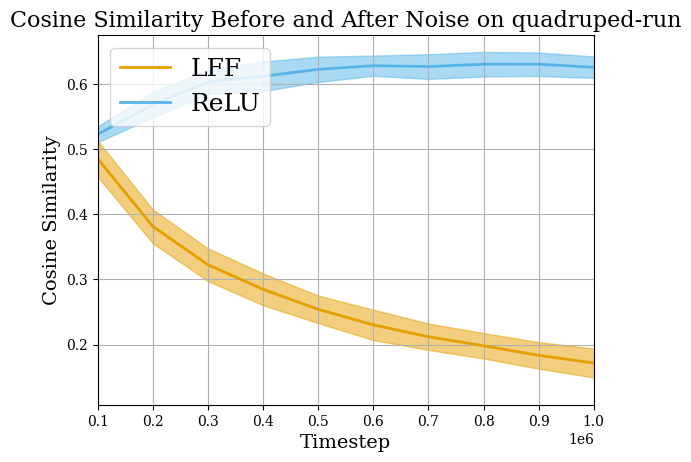}
    \caption{}
    \label{fig:cosine_similarity_activation}
\end{subfigure}
\begin{subfigure}{0.25\textwidth}
    \centering
    \includegraphics[width=.99\linewidth]{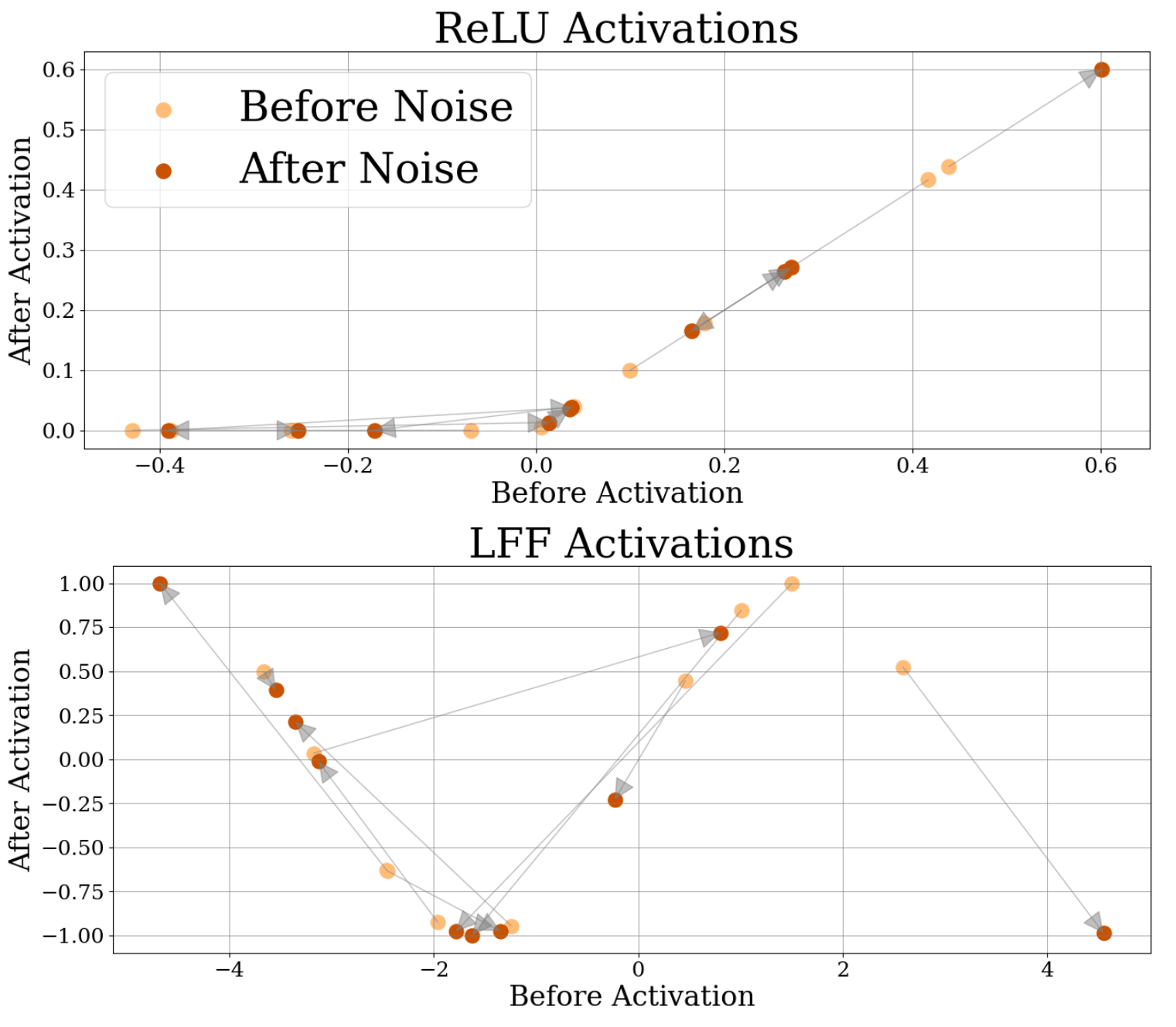}
    \caption{}
    \label{fig:activation_functions}
\end{subfigure}
\caption{Periodic activations are more brittle than ReLUs. (a) shows that $\sin$ pre and post-activations become less similar over time more quickly than ReLUs. (b) shows a similar trend in cosine similarity of activations before and after a medium level of noise is applied ($\sigma=0.625$). Qualitatively, LFF activations have their points shifted more dramatically than ReLU activations when inputs are perturbed (panel (c)). Results are reported across ten seeds. The shaded regions represent the standard deviation.}
\label{fig:diagnosing_generalization}
\end{figure}

\begin{tcolorbox}[colframe=black!50, colback=black!5, rounded corners]
Here we offer an explanation for the poor robustness to noise of learned Fourier features, showing that measures of network expressivity, such as effective rank \citet{kumar2020implicit}, are higher for Fourier representations than their equivalent ReLU representations.
\end{tcolorbox}
The relationship between the frequency of learned Fourier features and network generalization is perhaps best understood through the lens of feature geometry evolution. Prior work suggests measuring a neural network's expressivity by computing the cosine similarity between representations of two different inputs, formalized as the Q/C-maps of \citet{poole2016exponential}. To measure expressivity, we compute the similarity between $\mathbf{W}\mathbf{x}+\mathbf{b}$ and $g(\mathbf{W}\mathbf{x}+\mathbf{b})$ where $g$ is an activation function, plotting the result in figure \ref{fig:cosine_similarity_after_noise}---showing that ReLU representations see their
cosine similarity decrease more slowly than the Fourier features during training.

Layers that decorrelate their inputs (i.e. layers that reduce the cosine similarity before and after activations) tend to produce networks that are trainable but generalize poorly. On the other hand, functions that correlate their inputs on average allow networks to generalize more aggressively, potentially at the expense of convergence rates as discussed in greater detail by \citet{martens2021rapid}. As a further proxy measure of generalization, we compute the cosine similarity of representations before and after ``medium'' (see section 
\ref{sec:generalization}) noise perturbations. Explicitly, this is the similarity between $g(\mathbf{W}\mathbf{x} + \mathbf{b})$ and $g(\mathbf{W}(\mathbf{x} +\mathbf{\epsilon}) + \mathbf{b})$, where $\epsilon$ is a noise perturbation and $g$ is an activation function. We find that learned Fourier features more dramatically de-correlate inputs that differ by Gaussian noise (leading to lower similarities in figure \ref{fig:cosine_similarity_activation}).
\begin{table}
\centering
\begin{tabular}{lccc}
\toprule
\multicolumn{3}{c}{\textbf{Euclidean Distance of Representations Before and After Noise}} \\
\midrule
Environment & ReLU & LFF \\
\midrule
quadruped-walk & 74.00 $\pm$ 0.00 & 253.00 $\pm$ 0.00 \\
cheetah-run & 23.00 $\pm$ 0.00 & 224.40 $\pm$ 3.26 \\
acrobot-swingup & 7.20 $\pm$ 0.40 & 94.60 $\pm$ 3.07 \\
hopper-hop & 19.60 $\pm$ 0.49 & 211.40 $\pm$ 7.00 \\
walker-run & 29.00 $\pm$ 0.00 & 229.40 $\pm$ 1.36 \\
finger-turn\_hard & 15.00 $\pm$ 0.00 & 156.20 $\pm$ 6.85 \\
humanoid-run & 80.80 $\pm$ 0.75 & 253.00 $\pm$ 0.00 \\
quadruped-run & 74.00 $\pm$ 0.00 & 251.80 $\pm$ 0.40 \\
\midrule[0.15em]
Average & 40.33 $\pm$ 0.12 & 209.23 $\pm$ 1.36 \\
\bottomrule
\end{tabular}
\caption{Euclidean distance of representations before and after noise is added to inputs for ReLU and learned Fourier feature networks. On average, Learned Fourier feature representations are perturbed more than $5\times$ the distance that ReLU representations are when adding ``medium'' (see section 
\ref{sec:generalization}) Gaussian noise to inputs.}
\end{table}\label{tab:euclidean_distance_before_and_after_noise}
Additionally, we consider the Euclidean distance of activations before and after noise is added to the inputs. Euclidean distance before and after noise is added to the input is computed as $\|g(\mathbf{W}\mathbf{x}+\mathbf{b}) - (g(\mathbf{W}(\mathbf{x} + \epsilon)+\mathbf{b}))\|^{2}_{2}$. We hypothesize that small changes in the input observations will cause larger changes in the post-activation representations for Fourier features when compared to ReLU activations. We find this to be true empirically in table \ref{tab:euclidean_distance_before_and_after_noise}, where the Euclidean distance between post activation representations before and after noise is larger for learned Fourier feature representations than for ReLU representations. A qualitative visualisation of this is shown in figure \ref{fig:activation_functions}, where ReLU points are generally correlated before and after noise, while the Fourier representations are more brittle, with dramatic shifts before and after noise.

Lastly, recent works have quantified the expressivity of neural representations by computing their effective rank---including recent work on classical Fourier features that perform a fixed Fourier decomposition of states \cite{konidaris2011value, brellmann2023fourier}. Low-rank representations are ``implicitly underparameterized'' which may help generalization but decrease expressiveness \cite{kumar2020implicit, gulcehre2022empirical}. We compute the effective rank of representations from replay buffer samples at the end of training for both ReLU and learned Fourier feature representations in table \ref{tab:euclidean_distance_before_and_after_noise} showing the effective rank of Fourier features is many times larger than ReLU representations. 
\begin{table}
\centering
\begin{tabular}{lccc}
\toprule
\multicolumn{3}{c}{\textbf{Effective Rank}} \\

Environment & ReLU & LFF \\
\midrule
quadruped-walk & 23.90 $\pm$ 1.75 & 57.28 $\pm$ 0.34 \\
cheetah-run & 2.90 $\pm$ 0.16 & 9.65 $\pm$ 0.52 \\
acrobot-swingup & 0.64 $\pm$ 0.03 & 1.72 $\pm$ 0.04 \\
hopper-hop & 1.74 $\pm$ 0.39 & 8.69 $\pm$ 0.37 \\
walker-run & 6.96 $\pm$ 0.39 & 18.71 $\pm$ 0.31 \\
finger-turn\_hard & 2.18 $\pm$ 0.07 & 9.69 $\pm$ 0.75 \\
humanoid-run & 14.47 $\pm$ 2.30 & 50.84 $\pm$ 1.47 \\
quadruped-run & 24.35 $\pm$ 1.83 & 56.41 $\pm$ 0.31 \\
\midrule[0.15em]
Average & 9.64 $\pm$ 0.43 & 26.63 $\pm$ 0.23 \\
\bottomrule
\end{tabular}
\caption{Effective rank \citep{kumar2020implicit} of ReLU and learned Fourier features in Deepmind Control. Learned Fourier features have a higher effective rank than ReLU representations.}
\label{tab:rank_deepmind_control}
\end{table}
\subsection{Can the Generalization of Fourier Features be Improved?}\label{sec:fixing_generalization}
\begin{figure}
\centering
\begin{subfigure}{0.32\textwidth}
    \centering
    \includegraphics[width=.99\linewidth]{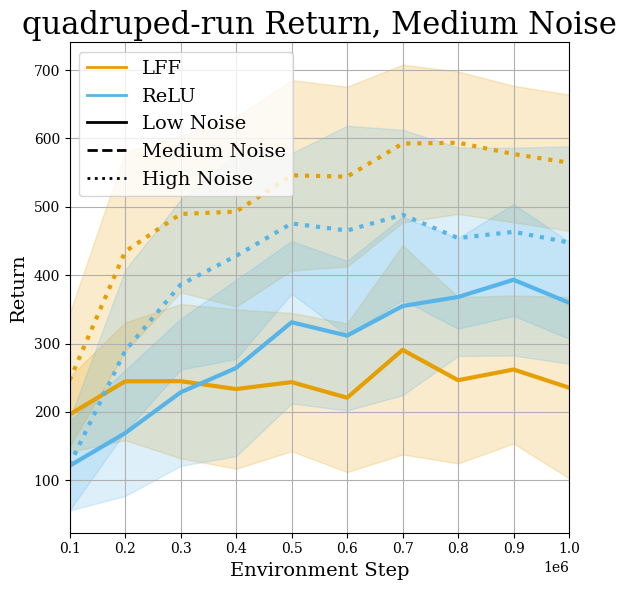}
    \caption{}
    \label{SUBFIGURE LABEL 2}
\end{subfigure}
\begin{subfigure}{0.3\textwidth}
    \centering
    \includegraphics[width=.99\linewidth]{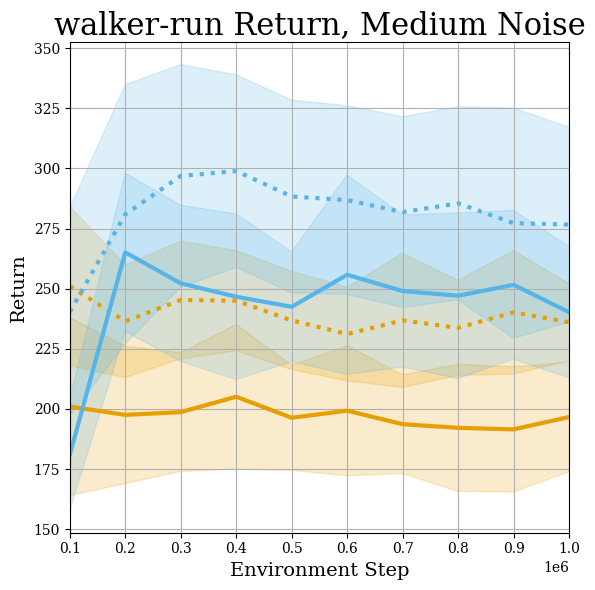}
    \caption{}
    \label{SUBFIGURE LABEL 3}
\end{subfigure}
\begin{subfigure}{0.3\textwidth}
    \centering
    \includegraphics[width=.99\textwidth]{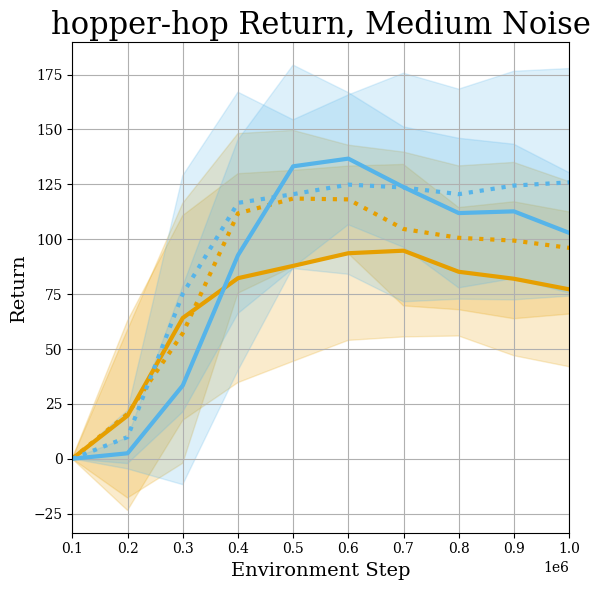}
    \caption{}
    \label{SUBFIGURE LABEL 1}
\end{subfigure}
\caption{Weight decay is able to partially offset overfitting at a medium level of observation noise, as introduced in section \ref{sec:fixing_generalization}, in some environments. In quadruped-run, weight decay allows learned Fourier features to generalize better than ReLU representations without weight decay, but does not improve upon ReLU in walker-run and hopper-hop. Results are reported across ten seeds. The shaded regions represent the standard deviation. Evaluation at different noise levels is performed at 100k increments throughout training.}\label{fig:weight_decay}
\end{figure}

\begin{tcolorbox}[colframe=black!50, colback=black!5, rounded corners]
The generalization performance of learned Fourier features can be improved with weight decay regularization, which punishes the growth of representation frequency.
\end{tcolorbox}
The previous sections suggest that learned Fourier representations fail to generalize because they learn high frequency representations. Here we introduce a weight decay term in the loss function, which effectively discourages frequencies from growing too large. For both ReLU and periodic activations, we find that a weight decay coefficient of 0.1 was optimal. In figure \ref{fig:weight_decay}, we plot how performance changes over training on the three environments from section \ref{sec:generalization}. We find the weight decay has a positive effect on generalization performance for both Fourier representations and ReLU representations, enabling learned Fourier features with weight decay to exhibit comparable or in some cases improved robustness to ReLU representations without weight decay. When aggregating results in the medium noise regime, LFF with weight decay is comparable to ReLU on 6/8 environments (acrobot-swingup, quadruped-run, hopper-hop, waker-run, cheetah-run and finger-turn\_hard) and reaches a higher return than ReLU on 2/8 environments in the low sample limit ($<300k$ environment steps for quadruped-run and for $<800k$ steps for humanoid-run, see figure \ref{fig:full_return_curves_medium} in the appendix). However, in all environments, ReLU with weight decay has improved or comparable return to LFF with weight decay, signaling that there is still a generalization gap to be closed between Fourier representations and ReLU representations. In supplementary, \label{app:full_general} we find that the effective rank of both ReLU and LFF weight decay representations is smaller, further suggesting that weight decay can improve the robustness of the representations evaluated.

\section{Conclusion}
We have shown that unconstrained learned Fourier features tend to converge to similar frequency representations on continuous control tasks regardless of their architecture. We show empirically that frequency learned by such architectures can be considered \textit{high frequency}---we test generalization by perturbing observation with noise as suggested by \citep{dulac2021challenges} and provide empirical explanations for why high frequency representations are brittle, showing that small changes in observations leads to large changes in Fourier representations. Lastly, we consider using weight decay as a means to discourage frequency from becoming excessively high---this is able to partially offset the generalization challenges faced by learned Fourier representations. An interesting avenue for future work would be to build adaptable architectures, that switch between low and high frequency representations depending on the novelty of incoming states.
\newpage
\newpage
\appendix
\section{Appendix}
\subsection{Further Results on Measuring Learned Fourier Frequencies}\label{app:measuring_frequencies}
\begin{figure}[H]
\centering
\begin{subfigure}{0.31\textwidth}
    \centering
    \includegraphics[width=.99\linewidth]{walker-run_scales_vs_return_CI.png}
    \caption{}
\end{subfigure}
\begin{subfigure}{0.31\textwidth}
    \centering
    \includegraphics[width=.99\textwidth]{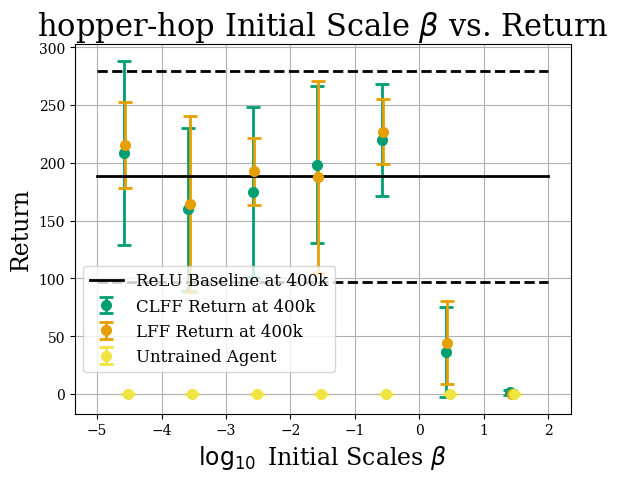}
    \caption{}
\end{subfigure}
\begin{subfigure}{0.32\textwidth}
    \centering
    \includegraphics[width=.99\linewidth]{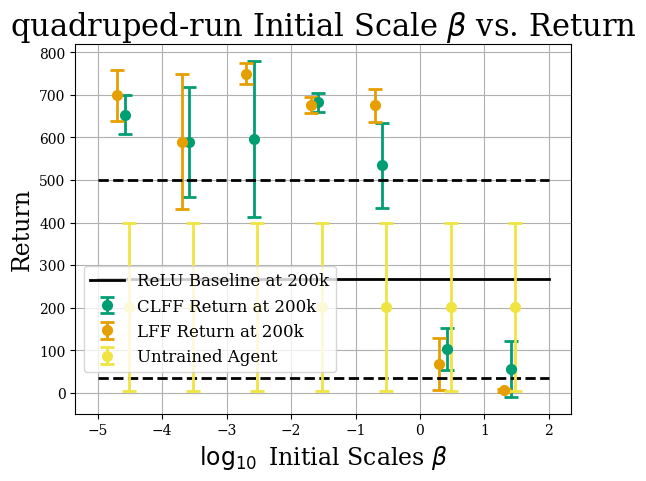}
    \caption{}
\end{subfigure}
\caption{Full sweep of different intialisations scales $\beta$ vs the return they receive early in training (100k for walker-run, 400k for hopper-hop and 200k for quadruped-run). On walker-run and quadruped-run Fourier features out perform the ReLU and untrained agents baselines.}
\end{figure}
\begin{figure}[H]
\centering
\begin{subfigure}{0.31\textwidth}
    \centering
    \includegraphics[width=.99\linewidth]{walker-run_scales_before_and_after_training_CI.png}
    \caption{}
\end{subfigure}
\begin{subfigure}{0.31\textwidth}
    \centering
    \includegraphics[width=.99\textwidth]{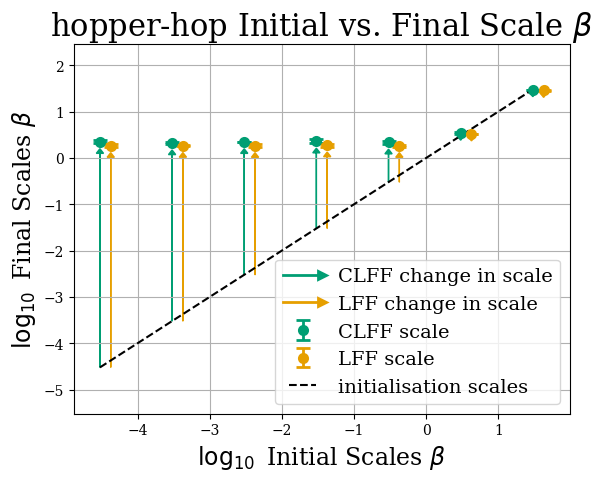}
    \caption{}
\end{subfigure}
\begin{subfigure}{0.32\textwidth}
    \centering
    \includegraphics[width=.99\linewidth]{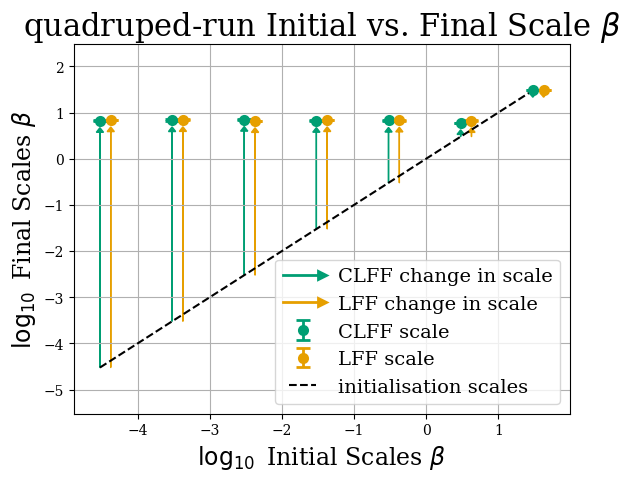}
    \caption{}
\end{subfigure}
\caption{Scales before and after training for the three different environments considered as well as for a sweep of different $\beta$ initializations. Provided that $\beta$ is not initialized too high, all learned Fourier feature representations tend to converge to similar frequencies.}
\end{figure}
\newpage
\newpage
\subsubsection{Full Deepmind Control Results for medium noise}\label{app:full_return_curves}
\begin{figure}[!htbp]
    \centering
    \includegraphics[width=.99\linewidth]{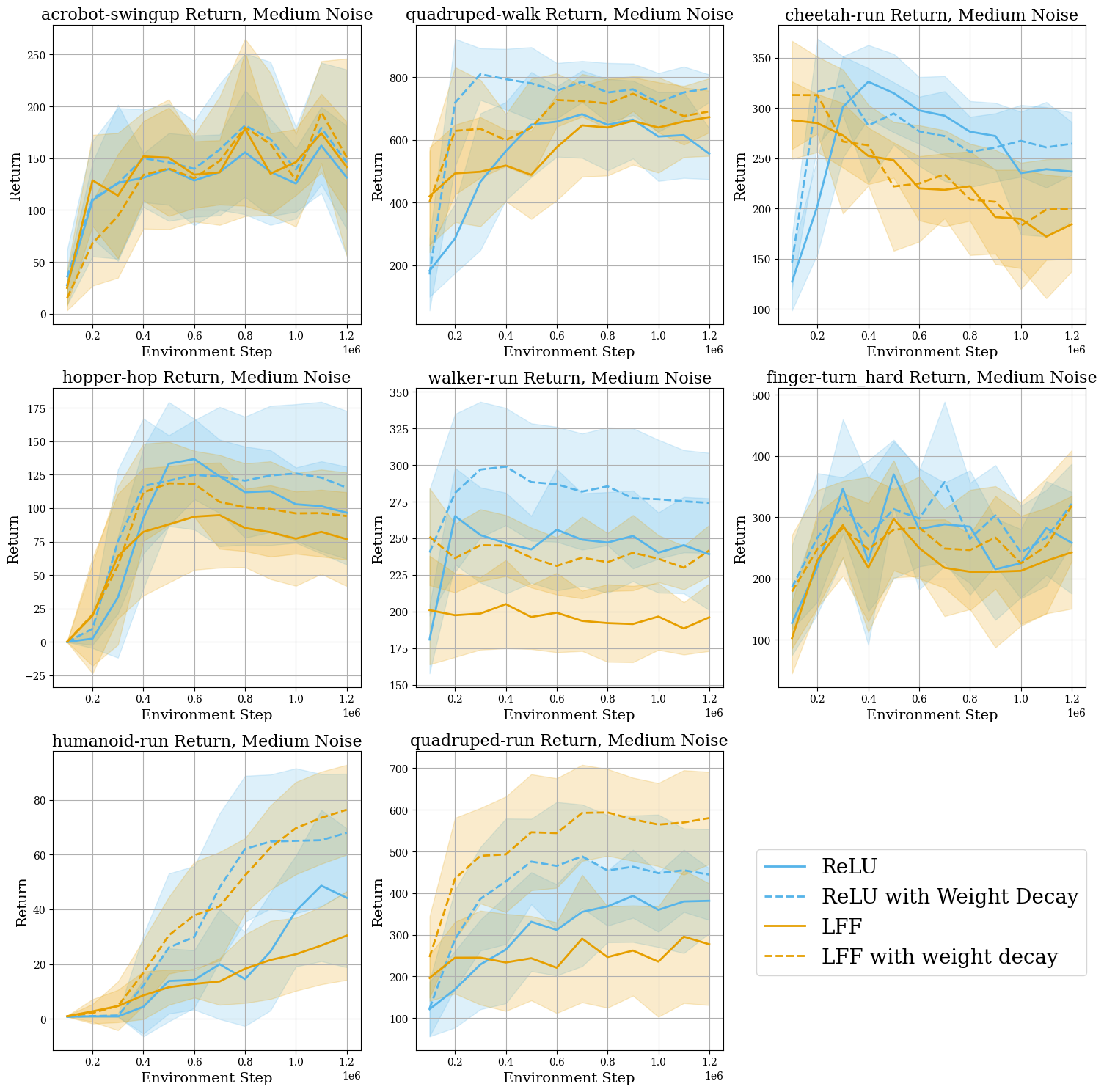}
    \caption{In the medium noise setting (see table \ref{tab:tuned_noise_levels_per_environment}) the learned Fourier features networks with weight decay outperform or match ReLU without weight decay on 6/8 environments (acrobot-swingup, quadruped-run, hopper-hop, waker-run, cheetah-run and finger-turn\_hard) and outperforms ReLU in the low sample limit in humanoid-run and quadruped-run.}
    \label{fig:full_return_curves_medium}
\end{figure}
\newpage
\bibliography{main}
\bibliographystyle{rlj}
\beginSupplementaryMaterials
\section{Further Results and Experimental Details}
\subsection{Supervised Learning Warmup}
\label{app:supervised_learning_warmup}
Following previous literature on learned Fourier features \cite{yang2022overcoming}, we qualitatively observe the performance of learned Fourier features on the mountain car environment \cite{moore1990efficient}. We first learn a ground truth value function with dynamic programming, using an implementation provided by \cite{yang2022overcoming}. Then we try to fit a MLP with ReLU activations and sinusoidal activations to the ground truth value function of the zeroth action, with a checkerboard pattern as the train/test split. We optimise both networks with Adam \cite{kingma2014adam} with a learning rate of $10^{-4}$. Each network has 2 hidden layers of width 400. The learned Fourier feature network has sinusoids throughout to highlight the properties of sinusoidal layers. We use an initial $\beta=10/\pi$. Note this experiment is used to exemplify the potential pitfalls of sinusoidal activations, it has not been tuned to achieve the best possible test error.
\subsection{Deepmind Control Hyperparameters}\label{app:deepmind_hyperparams}
Below we describe  in detail the different hyperparameters used by our learned Fourier features agents.
\subsubsection{Soft Actor-Critic hyperparameters}
\begin{table}[htbp]
\centering
\caption{SAC \cite{haarnoja2018soft} hyperparameters that are consistent throughout the baselines.}
\begin{tabular}{ll} 
\toprule
\textbf{Hyperparameter} & \textbf{Value} \\
\midrule
actor learning rate & $10^{-4}$ \\
critic learning rate & $10^{-4}$ \\
temperature learning rate & $10^{-4}$ \\
discount & 0.99 \\
target update period & 2 \\
initial temperature & 0.1 \\
batch size & 1024 \\
environment steps before training starts & $10^{4}$ \\
\bottomrule
\end{tabular}
\label{tab:example}
\end{table}
\subsubsection{ReLU architecture}\label{app:relu_architecture}
\begin{table}[!htbp]
\centering
\caption{ReLU Critic architecture}
\begin{tabular}{ll} 
\toprule
\textbf{Layer Output Dim} & \textbf{Activation} \\
\midrule
$20\times(\text{state dimension}+\text{action dimension})$ & \text{ReLU} \\
$1024$ & \text{ReLU} \\
$1024$ & \text{ReLU} \\
$1$ & \text{None} \\
\bottomrule
\end{tabular}
\label{tab:relu_critic_architecture}
\end{table}
\begin{table}[!htbp]
\centering
\caption{ReLU Actor architecture}
\begin{tabular}{l p{10cm}} 
\toprule
\textbf{Layer Output Dim} & \textbf{Activation} \\
\midrule
$1024$ & \text{ReLU} \\
$1024$ & \text{ReLU} \\
$2 \times \text{action dimension}$ & output is split into mean and log std dev that parametrizes \newline a Gaussian, tanh is applied to log std dev \\
\bottomrule
\end{tabular}
\label{tab:relu_actor_architecture}
\end{table}
\subsubsection{Learned Fourier Feature Architecture}
\begin{table}[htbp!]
\centering
\caption{Learned Fourier feature Critic architecture}
\begin{tabular}{ll} 
\toprule
\textbf{Layer Output Dim} & \textbf{Activation} \\
\midrule
$20\times(\text{state dimension}+\text{action dimension})$ & \text{Sinusoid} \\
$1024$ & \text{ReLU} \\
$1024$ & \text{ReLU} \\
$1$ & \text{None} \\
\bottomrule
\end{tabular}
\label{tab:LFF_critic_architecture}
\end{table}
The actor architecture is equivalent to the actor in table \ref{tab:relu_actor_architecture}.
\subsubsection{Concatenated Learned Fourier Feature Architecture}
\begin{table}[htbp!]
\centering
\caption{Concatenated Learned Fourier feature Critic architecture}
\begin{tabular}{ll} 
\toprule
\textbf{Layer Output Dim} & \textbf{Activation} \\
\midrule
$20\times(\text{state dimension}+\text{action dimension})$ & \text{Sinusoid, Cosine, None} \\
$1024$ & \text{ReLU} \\
$1024$ & \text{ReLU} \\
$1$ & \text{None} \\
\bottomrule
\end{tabular}
\label{tab:CLFF_critic_architecture}
\end{table}
Note the concatenated learned Fourier feature architecture uses a concatenation of a the inputs projected with a sinusoidal activation applied, the inputs projected with a cosine applied and then also just the raw inputs (see equation \ref{eqn:CLFF_initialisation}). The actor architecture is again equivalent to the actor in table \ref{tab:relu_actor_architecture}.
\subsection{Consistency with Reported Weight Standard Deviations from Li and Pathak 2021}\label{app:li_pathak_std_dev}
\citet{li2021functional} report a standard deviation of 0.05 for walker-run in figure 18 of their manuscript. To convert the standard deviation measurement to our $\beta$ measurement, we solve for $\beta$ in the following equation to convert between the conventions of describing weight initializations in \citet{li2021functional} and \citet{yang2022overcoming}:
\begin{equation}
    2\pi\sigma=\frac{\pi\beta}{d}
\end{equation}
Where as a reminder, $d$ in the size of the network input. In the case of walker-run $d=30$, meaning $\beta=2\pi d =2\times30\times0.05 = 3$, which is approximately equal to $2$, our $\beta$ measurement after training for walker-run.
\subsection{Tuning Observation Noise}\label{app:tuning_observation_noise}
We tuned observation noise for the different environments to reach low, medium and high levels of disruption on the performance of agents during training. Below is a table of the values of the standard deviation of the isotropic Gaussian noise added to each observations for the different environments.
\begin{table}[htbp!]
\centering
\caption{Tuned Levels of Observation Noise for Each Environment}
\begin{tabular}{lccc} 
\toprule
\textbf{Environment} & \textbf{Noise levels $\sigma$, [low, medium, high]} & action size & state size \\
\midrule
quadruped-walk & [0.125, 0.625, 1.0] & 12 & 58\\
cheetah-run & [0.125, 0.625, 1.0] & 6 & 17\\
acrobot-swingup & [0.05, 0.1, 0.5] & 1 & 6\\
hopper-hop & [0.125, 0.25, 0.5] &  4 & 15\\
walker-run & [0.25, 0.375, 1.0] &  6 & 24\\
finger-turn hard & [0.03, 0.1, 0.15] & 2 & 12\\
humanoid-run & [0.05, 0.25, 0.5] & 21 & 67  \\
quadruped-run & [0.25, 0.625, 1.0] & 12 & 58\\
\bottomrule
\end{tabular}
\label{tab:tuned_noise_levels_per_environment}
\end{table}
\subsection{Full Generalisation Properties results}\label{app:full_general}
\begin{table}[H]
\centering
\begin{tabular}{lcccc}
\toprule
Environment & ReLU & LFF & ReLU w/ weight decay & LFF w/ weight decay \\
\midrule
quadruped-walk & 23.90 $\pm$ 1.75 & 57.28 $\pm$ 0.34 & 5.65 $\pm$ 0.42 & 42.49 $\pm$ 1.85 \\
cheetah-run & 2.90 $\pm$ 0.16 & 9.65 $\pm$ 0.52 & 0.87 $\pm$ 0.05 & 4.60 $\pm$ 0.45 \\
acrobot-swingup & 0.64 $\pm$ 0.03 & 1.72 $\pm$ 0.04 & 0.16 $\pm$ 0.01 & 0.65 $\pm$ 0.06 \\
hopper-hop & 1.74 $\pm$ 0.39 & 8.69 $\pm$ 0.37 & 0.52 $\pm$ 0.03 & 3.81 $\pm$ 0.13 \\
walker-run & 6.96 $\pm$ 0.39 & 18.71 $\pm$ 0.31 & 1.88 $\pm$ 0.03 & 10.71 $\pm$ 0.23 \\
finger-turn\_hard & 2.18 $\pm$ 0.07 & 9.69 $\pm$ 0.75 & 0.54 $\pm$ 0.02 & 5.84 $\pm$ 0.40 \\
humanoid-run & 14.47 $\pm$ 2.30 & 50.84 $\pm$ 1.47 & 3.39 $\pm$ 0.73 & 44.21 $\pm$ 0.68 \\
quadruped-run & 24.35 $\pm$ 1.83 & 56.41 $\pm$ 0.31 & 6.02 $\pm$ 0.31 & 42.41 $\pm$ 0.98 \\
\midrule[0.15em]
Average & 9.64 $\pm$ 0.43 & 26.63 $\pm$ 0.23 & 2.38 $\pm$ 0.11 & 19.34 $\pm$ 0.29 \\
\bottomrule
\end{tabular}
\caption{Average effective rank over all environments, including the effective rank of weight decay representations}
\label{tab:relu_lff_comparison}
\end{table}
\begin{table}[H]
\centering
\begin{tabular}{lcccc}
\toprule
Environment & ReLU & LFF & ReLU w/ weight decay & LFF w/ weight decay \\
\midrule
quadruped-walk & 74.00 $\pm$ 0.00 & 253.00 $\pm$ 0.00 & 72.80 $\pm$ 0.40 & 250.40 $\pm$ 0.49 \\
cheetah-run & 23.00 $\pm$ 0.00 & 224.40 $\pm$ 3.26 & 23.00 $\pm$ 0.00 & 147.00 $\pm$ 12.60 \\
acrobot-swingup & 7.20 $\pm$ 0.40 & 94.60 $\pm$ 3.07 & 8.00 $\pm$ 0.00 & 51.20 $\pm$ 6.46 \\
hopper-hop & 19.60 $\pm$ 0.49 & 211.40 $\pm$ 7.00 & 19.00 $\pm$ 0.00 & 131.40 $\pm$ 5.95 \\
walker-run & 29.00 $\pm$ 0.00 & 229.40 $\pm$ 1.36 & 28.80 $\pm$ 0.40 & 193.60 $\pm$ 1.02 \\
finger-turn\_hard & 15.00 $\pm$ 0.00 & 156.20 $\pm$ 6.85 & 15.00 $\pm$ 0.00 & 72.60 $\pm$ 5.92 \\
humanoid-run & 80.80 $\pm$ 0.75 & 253.00 $\pm$ 0.00 & 79.80 $\pm$ 0.75 & 251.80 $\pm$ 0.40 \\
quadruped-run & 74.00 $\pm$ 0.00 & 251.80 $\pm$ 0.40 & 73.00 $\pm$ 0.00 & 249.20 $\pm$ 0.40 \\
\midrule[0.15em]
Average & 40.33 $\pm$ 0.12 & 209.23 $\pm$ 1.36 & 39.93 $\pm$ 0.12 & 168.40 $\pm$ 2.06 \\
\bottomrule
\end{tabular}
\caption{Average Euclidean distance before and after noise of the different representations considered (including weight decay options) for a medium level of noise.}
\label{tab:updated_methods_results}
\end{table}
\begin{figure}
\centering
\begin{subfigure}{0.31\textwidth}
    \centering
    \includegraphics[width=.99\linewidth]{walker-run_cycle_distribution_yang.png}
    \caption{}
    
\end{subfigure}
\begin{subfigure}{0.31\textwidth}
    \centering
    \includegraphics[width=.99\textwidth]{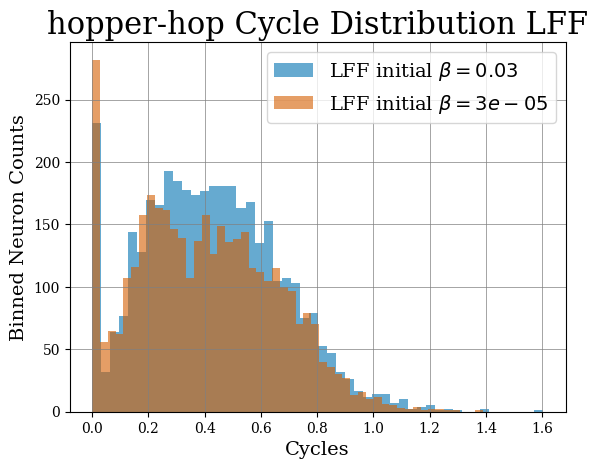}
    \caption{}
\end{subfigure}
\begin{subfigure}{0.32\textwidth}
    \centering
    \includegraphics[width=.99\linewidth]{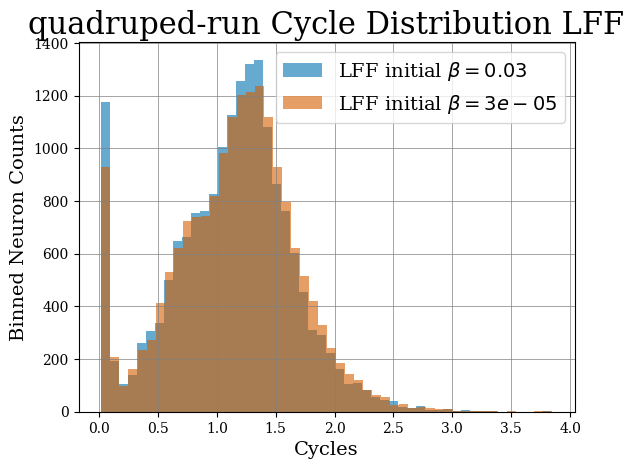}
    \caption{}
\end{subfigure}
\caption{Neuron frequencies (see definition \ref{def:representation_frequency} seem to not be influenced by initialization scale ($\beta$). On all 3 environments, the high $\beta$ and low $\beta$ histograms are mostly overlapping. Furthermore, all representations have a large high frequency component (bins with more than 0.25 cycles).}
\end{figure}
\begin{figure}
\centering
\begin{subfigure}{0.31\textwidth}
    \centering
    \includegraphics[width=.99\linewidth]{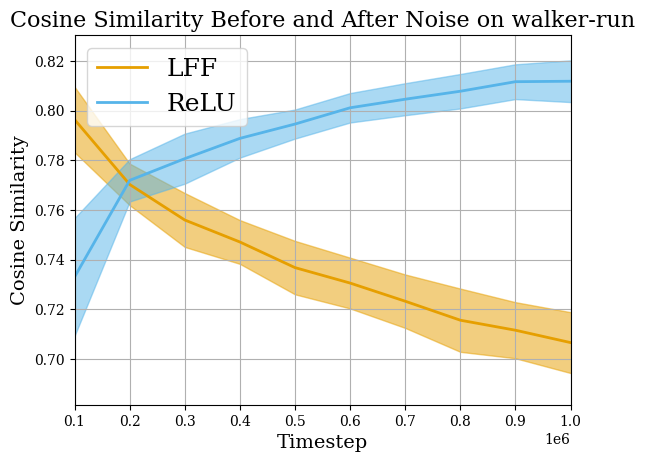}
    \caption{}
\end{subfigure}
\begin{subfigure}{0.31\textwidth}
    \centering
    \includegraphics[width=.99\textwidth]{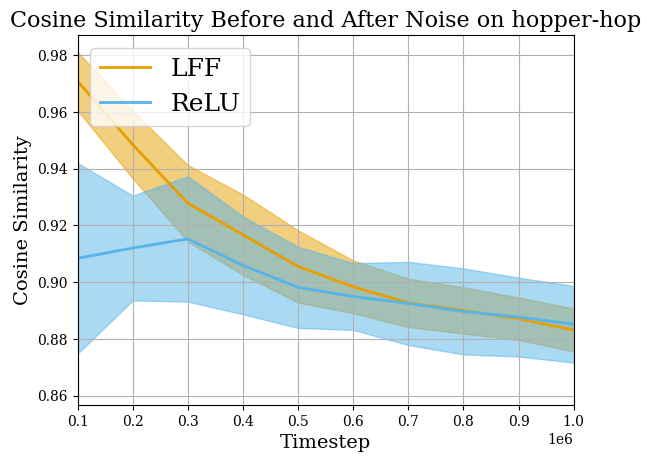}
    \caption{}
\end{subfigure}
\begin{subfigure}{0.32\textwidth}
    \centering
    \includegraphics[width=.99\linewidth]{cosine_similarity_quadruped-run_before_and_after_noise.png}
    \caption{}
\end{subfigure}
\caption{Cosine similarity before and after a medium level of noise for all three environments.}
\end{figure}
\begin{figure}
\centering
\begin{subfigure}{0.31\textwidth}
    \centering
    \includegraphics[width=.99\linewidth]{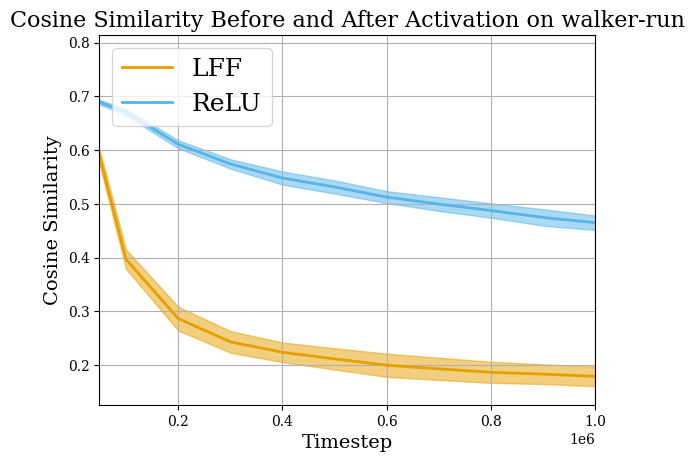}
    \caption{}
\end{subfigure}
\begin{subfigure}{0.31\textwidth}
    \centering
    \includegraphics[width=.99\textwidth]{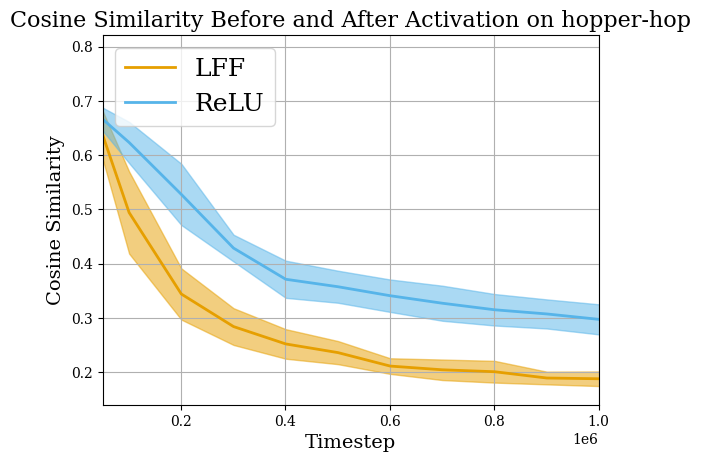}
    \caption{}
\end{subfigure}
\begin{subfigure}{0.32\textwidth}
    \centering
    \includegraphics[width=.99\linewidth]{cosine_similarity_quadruped-run_before_and_after_activation.png}
    \caption{}
\end{subfigure}
\caption{The cosine similarity before and after activations for the three environments considered.}
\end{figure}
We plot the results for all environments for no noise training curves and evaluation curves for low, medium and high levels of noise (as described in table \ref{tab:tuned_noise_levels_per_environment}).
\begin{figure}[!htbp]
    \centering
    \includegraphics[width=.99\linewidth]{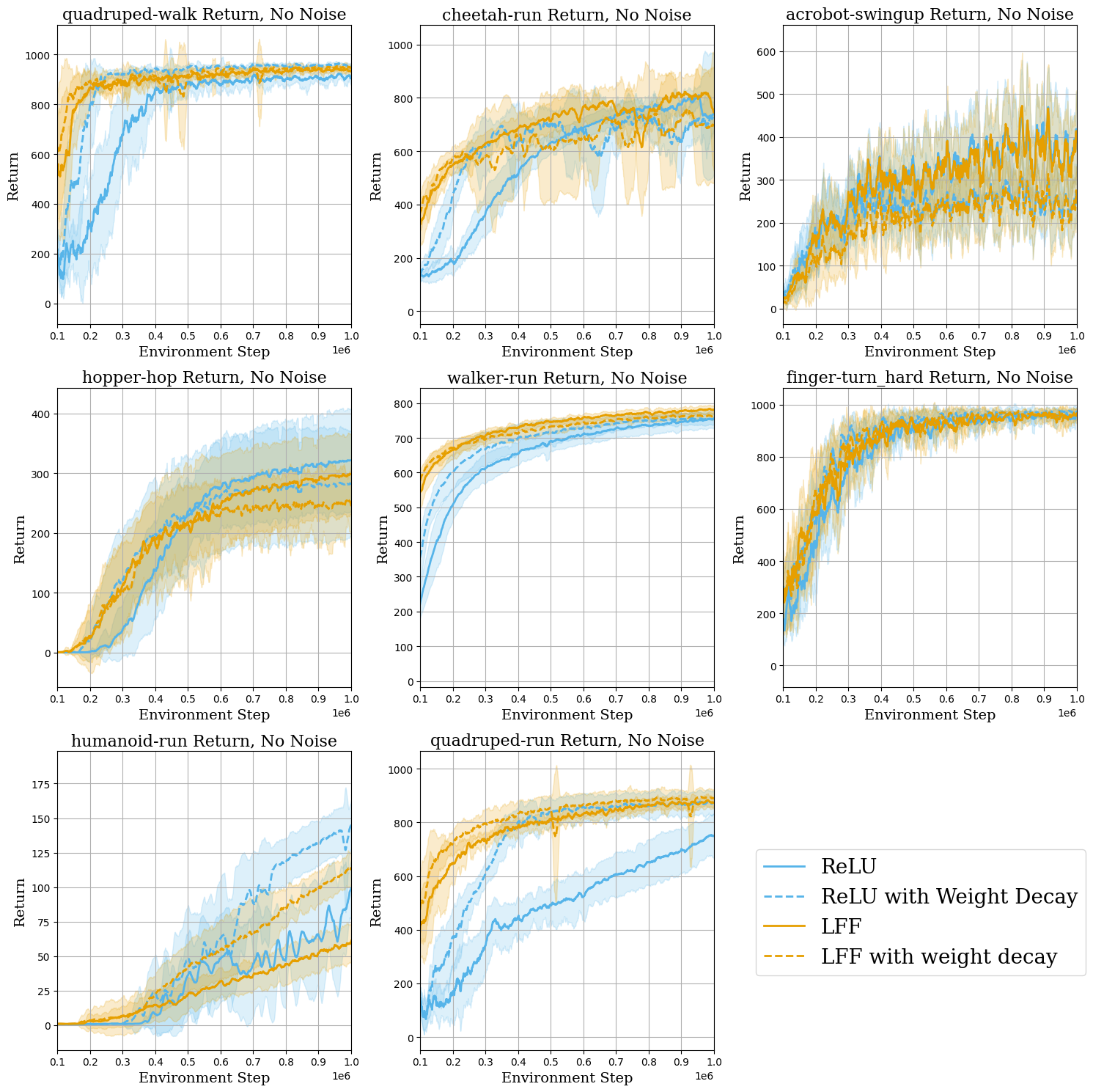}
    \caption{Training curves across the full Deepmind control suite, with and withou decay for learned Fourier features, concatenated learned Fourier features and ReLU activations.}
    \label{fig:no_noise_return_curves}
\end{figure}
\begin{figure}[!htbp]
    \centering
    \includegraphics[width=.99\linewidth]{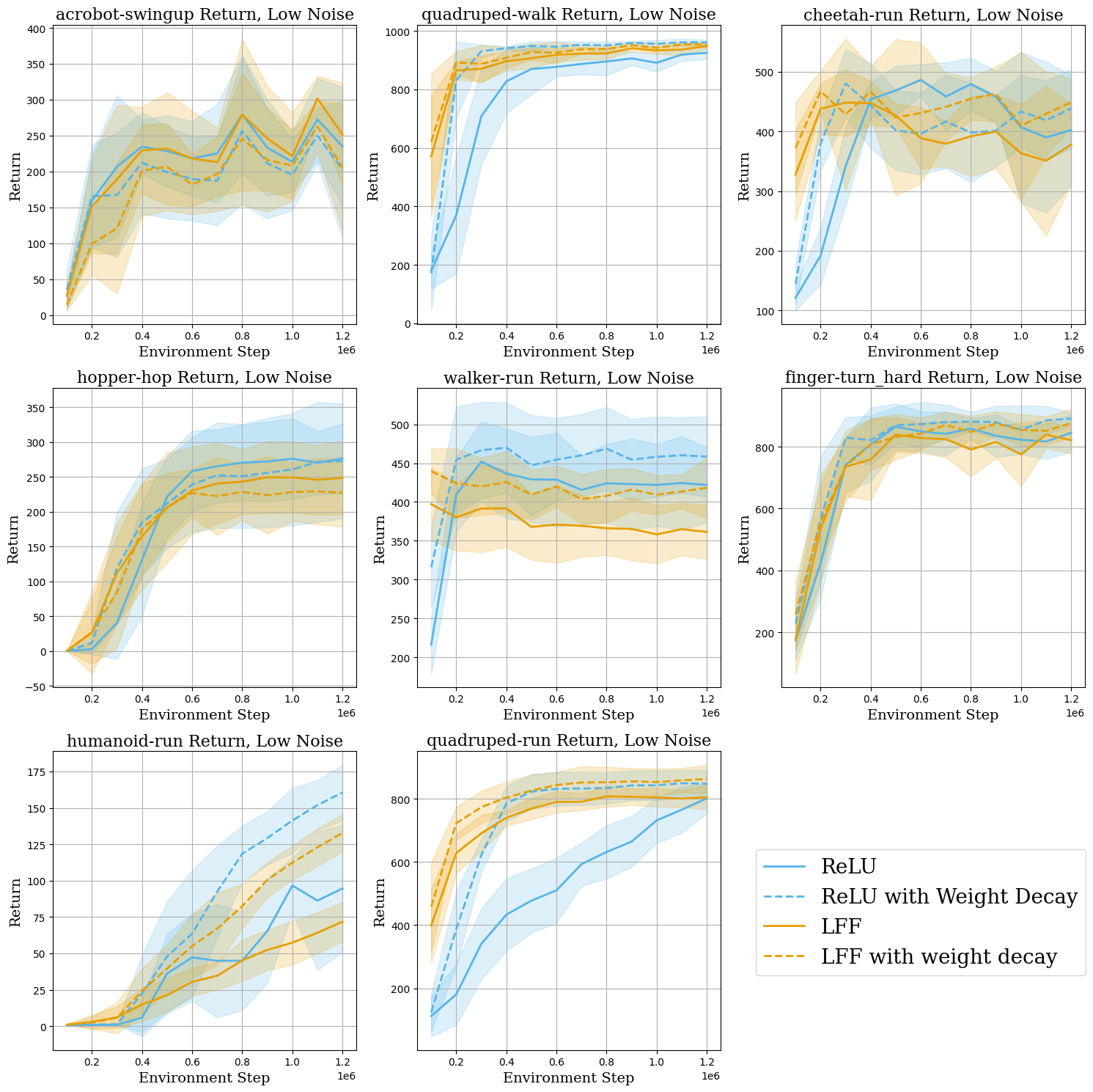}
    \caption{Full evaluating curves for all 8 environments plotted for a low amount of noise (see table \ref{tab:tuned_noise_levels_per_environment}). Here the benefits of ReLU activations over LFF representations are still present in one environment (quadruped-walk), however even with this low level of noise ReLU activations begin to outperform or match Fourier feature representations.}
    \label{fig:full_return_curves}
\end{figure}
\begin{figure}[!htbp]
    \centering
    \includegraphics[width=.99\linewidth]{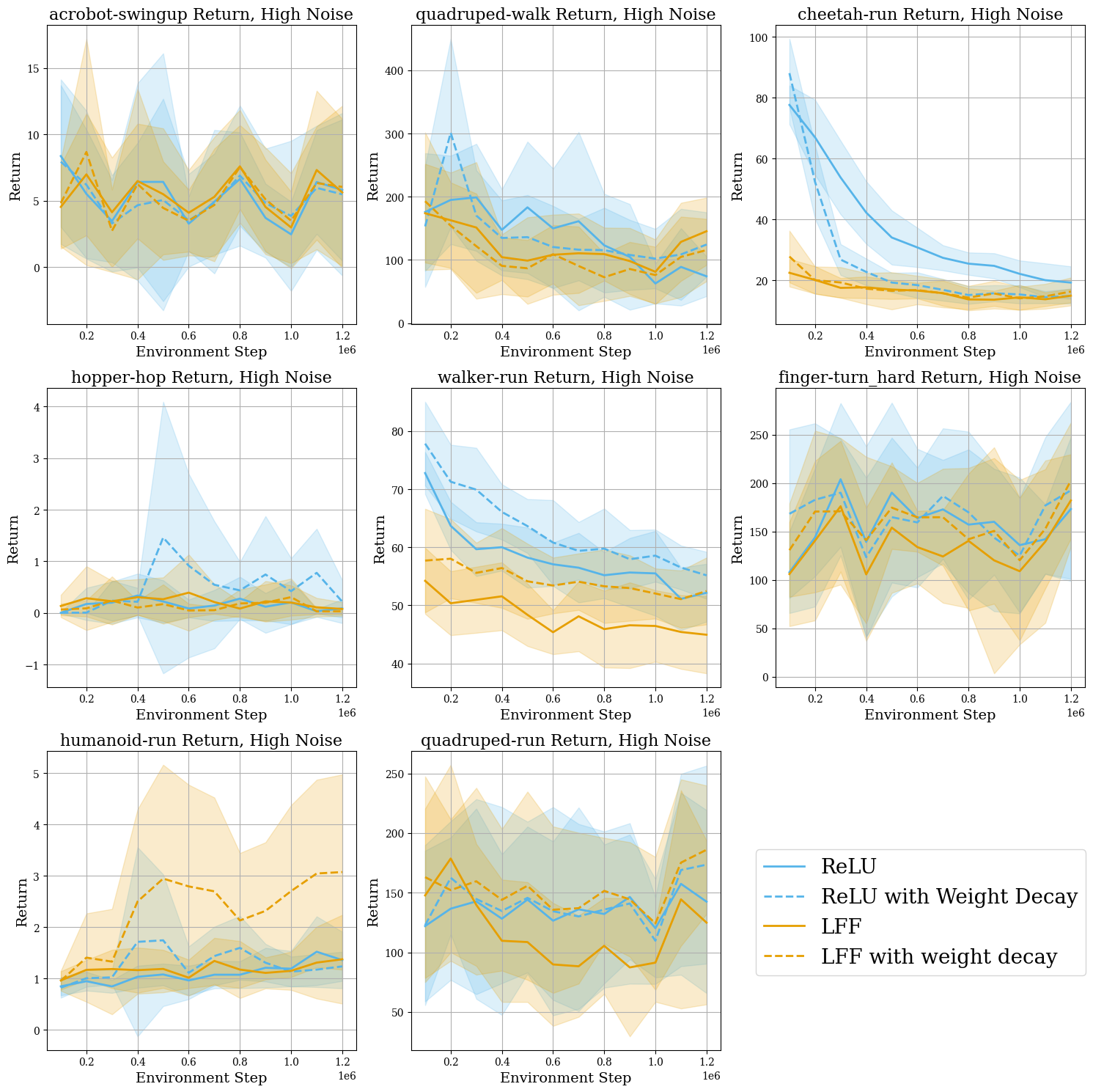}
    \caption{In the high noise regime, all algorithms exhibit poor performance on all 8 environments.}
    \label{fig:full_return_curves}
\end{figure}
\end{document}